\documentclass[runningheads]{llncs}
\usepackage{graphicx}
\usepackage{booktabs}
\usepackage{todonotes}
\usepackage{url}

\begin{document}
\title{A Dataset and Strong Baselines for Classification of Czech News Texts}
\author{Hynek Kydlíček \and Jindřich Libovický\orcidID{0000-0001-7717-4090}
 }
%
\authorrunning{H. Kydlíček \& J. Libovický}
%
\institute{Charles University, Faculty of Mathematics and Physics, Institute of Formal and Applied Linguistics, Malostranské nám. 25, 118 00 Prague, Czech Republic
\email{kydlicek.hynek@gmail.com, libovicky@ufal.mff.cuni.cz}}
\maketitle              
\begin{abstract}
Pre-trained models for Czech Natural Language Processing are often evaluated on purely linguistic tasks (POS tagging, parsing, NER) and relatively simple classification tasks such as sentiment classification or article classification from a single news source. As an alternative, we present CZEch~NEws~Classification~dataset (CZE-NEC), one of the largest Czech classification datasets, composed of news articles from various sources spanning over twenty years, which allows a more rigorous evaluation of such models. We define four classification tasks: news source, news category, inferred author's gender, and day of the week. To verify the task difficulty, we conducted a human evaluation, which revealed that human performance lags behind strong machine-learning baselines built upon pre-trained transformer models. Furthermore, we show that language-specific pre-trained encoder analysis outperforms selected commercially available large-scale generative language models.

\end{abstract}

\section{Introduction}

Natural Language Processing (NLP) tools in Czech are often evaluated on purely linguistic tasks such as POS tagging, dependency parsing from Universal Dependencies \cite{nivre-etal-2017-universal}, or Named Entity Recognition (NER) \cite{ner}.  As linguistic tools lose importance as parts of more complex NLP pipelines, semantic and pragmatic end-to-end tasks become more important evaluation benchmarks. The downstream tasks available for Czech include sentiment analysis \cite{habernal-etal-2013-sentiment}, news topic classification \cite{kral-lenc-2018-czech}, or text summarization \cite{strakaSumeCzechLargeCzech2018a}. Compared to other languages, the number of interesting NLP tasks is limited.

With large language models being able to operate multilingually, there is a new need for challenging evaluation datasets beyond English. Most NLP tasks also work with short texts, even though longer texts pose a bigger challenge for Transformer-based models, which rely on self-attention with quadratic memory complexity. We fill this gap by introducing a new dataset with challenging tasks for both machine learning models and humans.

We create the CZE-NEC by crawling Czech news websites from CommonCrawl (\S~\ref{sec:dataset_creation}) and use the available metadata to define classification tasks (\S~\ref{sec:task_definition}). The tasks are: news source classification, news category classification, inferred gender of the author (to assess a risk of gender discrimination based solely on text authorship), and day of the week when the news was published. We estimate the actual difficulty of the tasks by the human performance measured on a sample of the test data (\S~\ref{sec:human_annotation}).

Finally, we present strong baselines for the dataset (\S~\ref{sec:experiments}) using state-of-the-art machine learning models. Recently, several pre-trained encoder-only models for Czech were introduced \cite{strakaRobeCzechCzechRoBERTa2021,leheckaComparisonCzechTransformers2021} that reach state-of-the-art results both on existing benchmarks and our dataset. They outperform estimated human performance on all tasks, and on two tasks (cases), they outperform fine-tuned GPT-3 model \cite{gpt3}. 

\section{CZEch NEws Classification dataset (CZE-NEC)}

CZE-NEC is compiled from news stories published online in major Czech media outlets between January 2000 and August 2022. The news article content is protected by copyright law; therefore, we cannot distribute the dataset directly. Instead, we release software~\footnote{\url{https://github.com/hynky1999/Czech-News-Classification-dataset}} for collecting the dataset.

\subsection{Dataset Creation Process}\label{sec:dataset_creation} 
We have collected the news stories text from the following six Czech online news providers: \emph{SeznamZprávy.cz}, \emph{iRozhlas.cz}, \emph{Novinky.cz}, \emph{Deník.cz}, \emph{iDnes.cz}, and \emph{Aktuálně.cz}. Instead of crawling the pages directly, we used the CommonCrawl archive to extract the articles.

\subsubsection{Filtering.}
Not all pages on the news websites are news articles. Pages may contain videos, photo galleries, or quizzes, i.e., typically JavaScript code, which needs to be filtered out. We applied common data-cleaning techniques, namely language identification and rule-based filtering.  We used the FastText Language detection~\cite{joulin2016bag,joulin2016fasttext} model to filter out non-Czech articles, requiring all lines to be classified as Czech. To remove wrongly parsed articles, we kept only ones with the following properties: content length of at least 400 characters, headline length of at least 20 characters, and brief length of at least 40 characters. To exclude content that is not text, we only kept articles with the following properties:

\begin{enumerate}
\item The average word length is at least 4;
\item The number of words per total article length in characters is in the interval $(0.11, 0.22)$; and
\item The ratio of non-alphanumeric characters is at most $4.5\%$ per Length - $(0, 0.045)$.
\end{enumerate}

We also dropped articles with prefixes indicating non-news content, such as video, photo, or gallery. Finally, we removed articles with identical briefs, headlines, or content.

\subsubsection{Dataset Postprocessing.}
After filtering, we manually merged similar categories, resulting in 25 final categories, and filtered authors to 11k unique ones. We then removed excluded labels from the dataset. Content, brief, and headline were post-processed, including Unicode and HTML normalization and formatting adjustments. Gender was inferred from the authors using Namsor\footnote{https://namsor.app/}. If the article contained more than one author, we chose the homogeneous gender if possible. Otherwise, we labeled the Gender as Mixed. Even though the estimation provided by Namsor is likely to be correct in most cases, we realize that the actual gender cannot be inferred from a name. Individuals can identify with different gender that does not correspond to the linguistic features of their name.  We discuss this issue also later in the paper. 

\subsubsection{Splits.}\label{sec:splits}
\begin{figure}[t]
    \centering
    \includegraphics[width=1.0\textwidth]{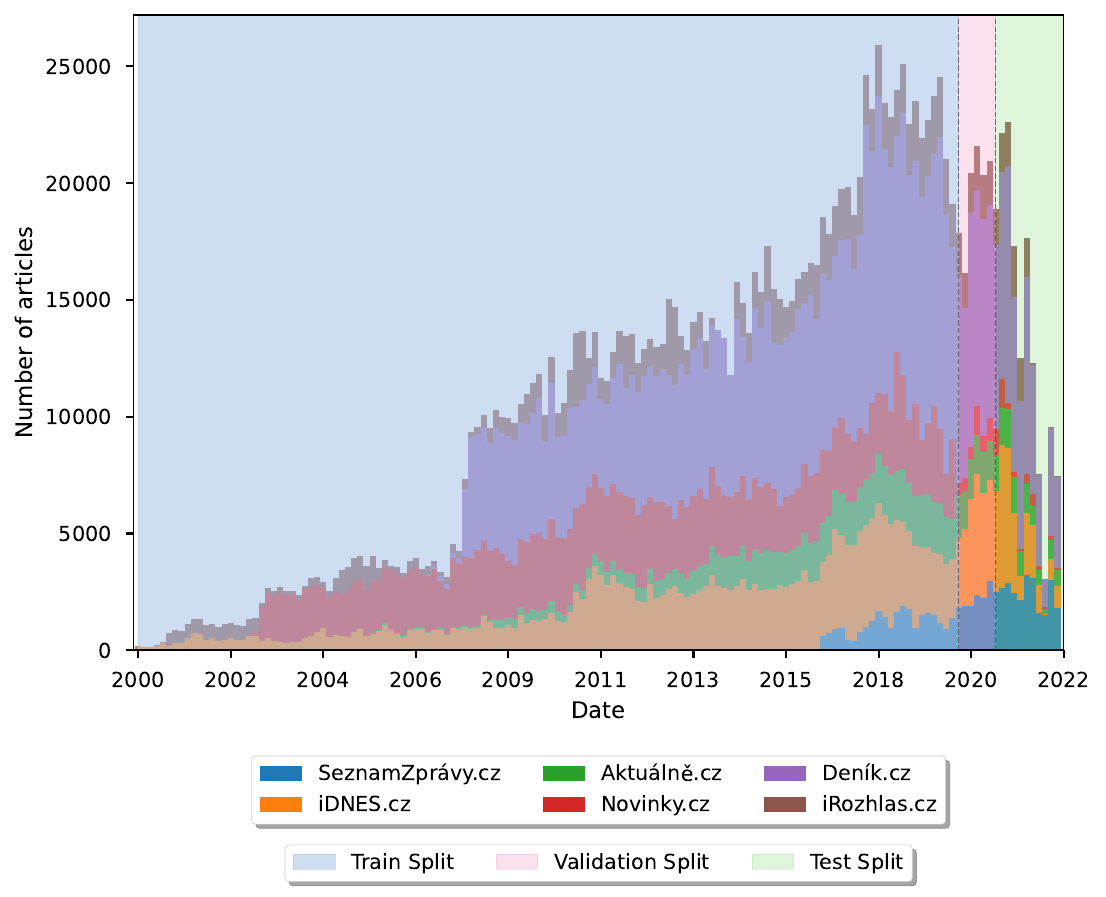}
    \caption{Distribution of news sources over time with dataset split boundaries.}
    \label{fig:dst_dist}
\end{figure}

We divided the dataset into the train, validation, and test sets based on publication date, using a 34:3:3 ratio. The splits are chronological, i.e., articles in the training set were published \emph{before} the test set, so we can asses if the models generalize in time. The dataset division is depicted in Figure~\ref{fig:dst_dist}.

\subsection{Dataset Summary}

\begin{table}[t]
        \caption{Dataset summary. Article words were calculated based
            on Moses tokenization.}
        \label{tab:dataset_summary}
        \centering
        \begin{tabular}{l ccccc}
            \toprule
            Source          & Size    & Authors & Categories & Start date            & Words per article \\
            \midrule
            Deník.cz   & \hphantom{1,}664,133  & \hphantom{1}2,497           & 18         & 2007  & 332           \\
            Novinky.cz   & \hphantom{1,}321,417  & \hphantom{1}2,518           & 17         & 2002 & 274           \\
            iDnes.cz & \hphantom{1,}295,840  & \hphantom{1}4,386           & 21         & 2000   & 423           \\
            iRozhlas.cz   & \hphantom{1,}167,588  & \hphantom{1}1,900           & 8          & 2000  & 287           \\
            Aktuálně.cz   & \hphantom{1,}112,960  & \hphantom{10,}633            & 19         & 2005 & 468           \\
            SeznamZprávy.cz~~~ & \hphantom{1,0}65,472   & \hphantom{10,}382            & 11         & 2016  & 443           \\
            \midrule
            Total       & 1,627,410 & 10,930          & 25         & 2000   & 362           \\
            \bottomrule
            \\
        \end{tabular}
\end{table}

The dataset contains the following data items for each article:
\begin{itemize}
    \item \emph{Source} -- Website that published the article;
    \item \emph{Content} -- Actual text content of the article;
    \item \emph{Brief} -- Brief/Perex of the article;
    \item \emph{Headline} -- Headline/Title of the article;
    \item \emph{Category} -- Both post-processed and original category;
    \item \emph{Published Date} -- Date of publication and inferred day of the week;
    \item \emph{Inferred gender} -- Inferred gender of author(s) name(s);
    \item \emph{Keywords} -- Extracted keywords from the article; and
    \item \emph{Comments Count} -- Number of comments in the discussion section.
\end{itemize}
Basic statistics summarizing the dataset are presented in Table \ref{tab:dataset_summary}.

\subsection{Task Definitions}\label{sec:task_definition}
\begin{table}[t]
        \caption{Tasks distribution over sets.}
        \label{tab:task_distribution}
        \centering
        \begin{tabular}{l cccc}
        \toprule
        Set &   Source &  Category &  Gender &  Day of week \\
        \midrule
        Train      &  1,383,298 &             \hphantom{1,}879,019 & \hphantom{1,}919,840 &      1,383,298 \\
        Validation~~~ &  \hphantom{1,}122,056 &  \hphantom{1,0}78,084 & \hphantom{1,0}82,936 & \hphantom{1,}122,056 \\
        Test       &  \hphantom{1,}122,056 &  \hphantom{1,0}82,352 & \hphantom{1,0}83,269 & \hphantom{1,}122,056 \\
        \midrule
        Total   &   1,627,410 &  1,039,455 &           1,086,045 &    1,627,410  \\
        \bottomrule
        \end{tabular}
\end{table}

The input for each task is only the article body, without including the brief or the headline. Not all articles contain all metadata; not all are available for all tasks. The distribution of samples across tasks is in Table~\ref{tab:task_distribution}. In the following paragraphs, we describe the four classification tasks in more detail.

\subsubsection{Source.}
\label{sec:server-desc}
\begin{figure}[h]
    \centering
    \includegraphics[width=1.0\textwidth]{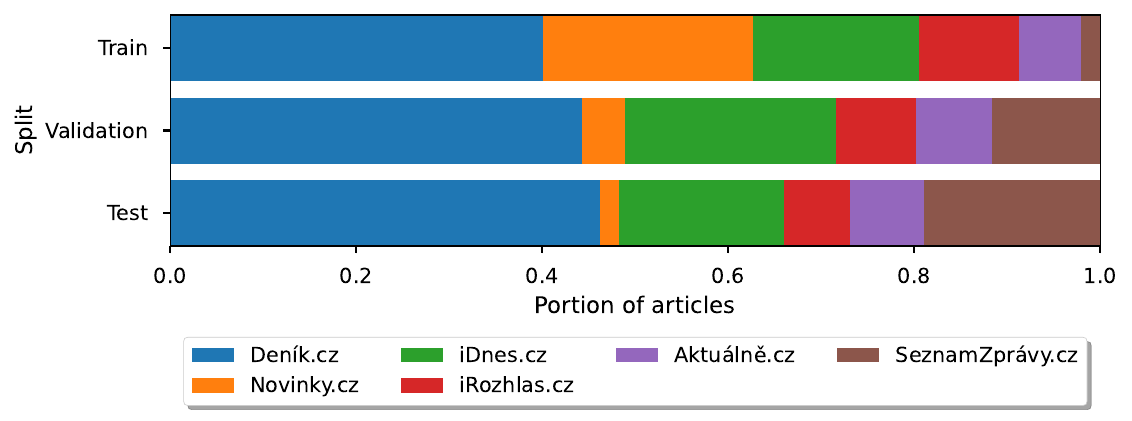}
    \caption{Dataset's label distribution of the Source task.}
    \label{fig:server_graph}
\end{figure}
The source classification task involves predicting the publishing website of articles from a set of 6 labels, as shown in Figure~\ref{fig:server_graph}. It is important to note that there is a significant distribution shift between the training and validation set, which is caused by differences in the launch dates of the websites and parsing issues (especially with Novinky.cz).

\subsubsection{Category.}
\begin{figure}[ht]
    \centering
    \includegraphics[width=1.0\textwidth]{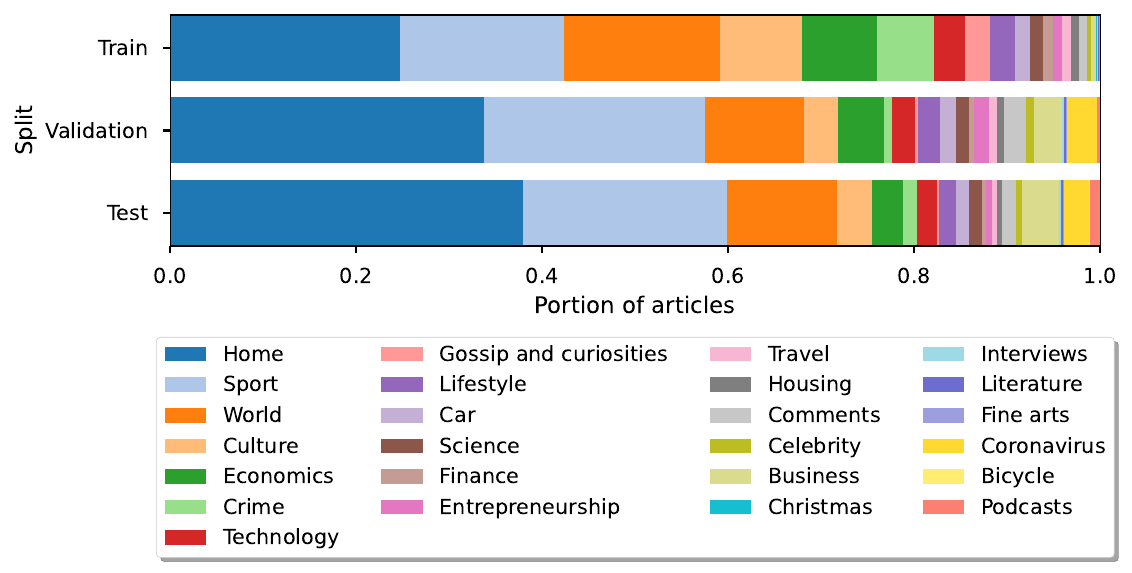}
    \caption{Dataset's label distribution of the Category task.}
    \label{fig:category_graph}
\end{figure}
The Category classification task requires predicting the category of an article from a set of 25 labels, as depicted in Figure~\ref{fig:category_graph}. When selecting the categories, we carefully identify the most frequent ones while striving to maintain diversity and minimize any potential overlap between them. 

We acknowledge that certain category selections might be disputed in some cases. For instance, we could have merged more similar categories, such as Entrepreneurship and Business. Likewise, some categories should have been separate such as Lifestyle and Health. Lastly, we should have considered excluding the Home and Foreign sections due to their span. Therefore we decided also to include the original categories without merges in CZE-NEC.

\subsubsection{Inferred gender.}
\begin{figure}[ht]
    \centering
    \includegraphics[width=1.0\textwidth]{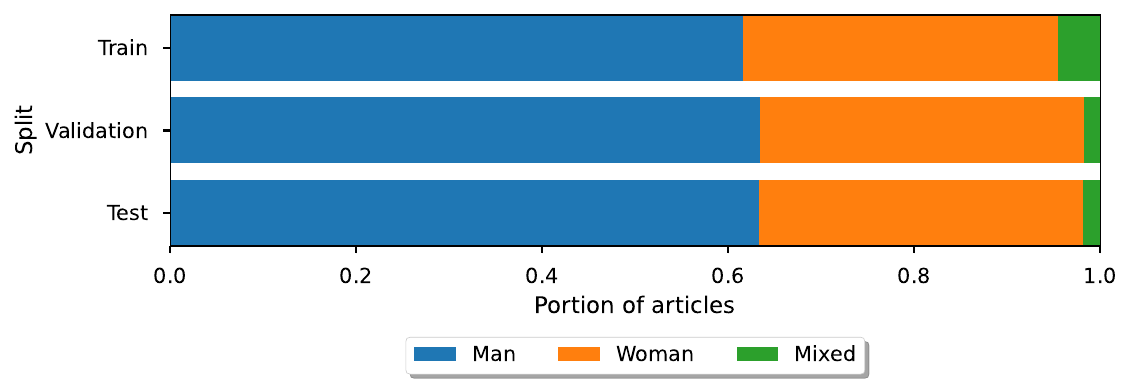}
    \caption{Dataset's label distribution of the Gender task.}
    \label{fig:gender_graph}
\end{figure}
This classification task has 3 labels, as shown in Figure~\ref{fig:gender_graph}. We acknowledge that accurately inferring the gender from a person's name is, in principle, impossible, as the actual gender might not correspond to the linguistic features of the names. Inaccurately assuming gender is potentially harmful to individuals whose names are being labeled. Unlike other languages, Czech has a stronger association of social and grammatical gender than many other languages, which makes inferring gender more accurate. We thus consider the inferred gender for names to be a reasonable proxy for our purposes. The goal of the task is to find if neural models consider authors' gender, which could potentially lead to discriminatory output in other NLP tasks. We only work with accumulated approximate statistics, which we believe are a reasonable approximation of the social reality. This task is not meant to label individuals and the text they produce, and we discourage future users of CZE-NEC from doing so. There also might be cases (especially reports taken from news agencies) where the author signed under the paper might not be the main author of the text. Given the approximate nature of the inferred gender classification, we believe it does not influence the meaningfulness of the task.

\subsubsection{Day of the Week.}
\begin{figure}[h]
    \centering
    \includegraphics[width=1.0\textwidth]{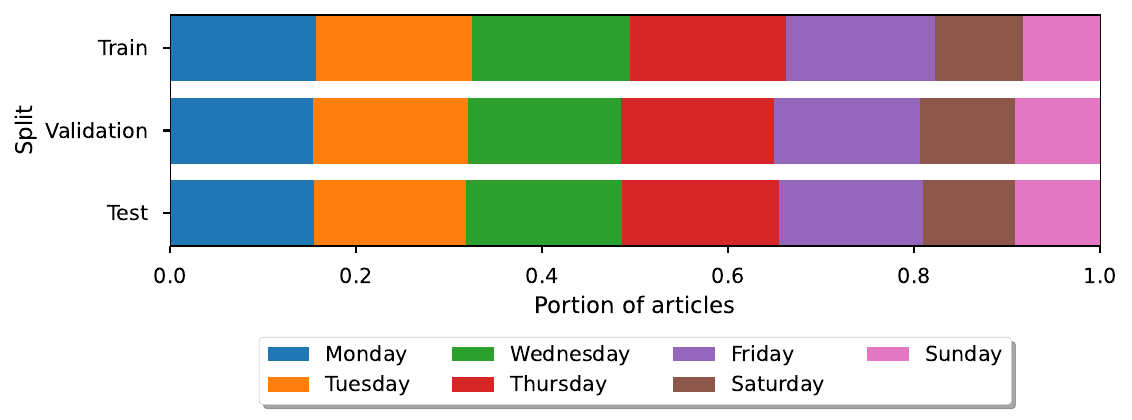}
    \caption{Dataset's label distribution of Day Of Week task.}
    \label{fig:day_graph}
\end{figure}
The Day of Week task is a classification challenge consisting of seven distinct labels, as illustrated in Figure~\ref{fig:category_graph}. The objective is to accurately predict the day of the week a given article was published. Given the absence of any apparent approaches to tackle this task, we deem it to be the most challenging among the tasks considered.

\subsection{Human Annotation}\label{sec:human_annotation}
To assess the human performance on the tasks, four instructors were instructed to assign labels for each article. We used a smaller dataset of 100 randomly selected samples from the test set, encompassing all associated metadata~(\textbf{Test Human}).

Upon completing the annotation process, the scores from each evaluator were averaged to derive the \textbf{Human} score. The human scores are presented along with the model results in Table~\ref{tab:results-human}.

\section{Task Baselines}\label{sec:experiments}

We present baseline experiments in two data regimes. In the first one, we use the entire CZE-NEC and present results using state-of-the-art pre-trained encoders. We anticipate that less training data will suffice with larger and more pre-trained language models. Therefore, we include a second, smaller data setup and GPT-3 among the baselines.

\subsection{Large Dataset Setup}
In this scenario, we train and evaluate various models on the unrestricted train/test sets. First, we employ logistic regression to assess the performance of a keywords-based model. A possible high accuracy of this model would mean that the tasks are solvable only by spotting typical keywords for the classes without any deeper understanding. Subsequently, we fine-tune two Czech pre-trained Transformer encoders for our tasks. We do not test GPT-3 in this setup,  mostly due to the high costs of such finetuning.

\subsubsection{Logistic Regression.}
We used Logistic Regression (LR) for the baseline model with TF-IDF and several additional features.
Following \cite{strakaSumeCzechLargeCzech2018a}, we incorporated the following features:
\begin{itemize}
    \item Number of words;
    \item Number of words with only non-alphabetic characters;
    \item Number of uppercase words;
    \item Number of digits words; and
    \item Number of capitalized words.
\end{itemize} We used 1-2 grams with features max document frequency: 1\% and restricted total TF-IDF features to 200k. We use logistic regression as our first baseline to rule out that the task we define is too simple and can be solved using keyword spotting.

\subsubsection{Fine-tuning pre-trained encoders.}\label{sec:base}
We experimented with two Czech transformer models, \textbf{RobeCzech}~\cite{strakaRobeCzechCzechRoBERTa2021} and \textbf{Fernet-News}~\cite{leheckaComparisonCzechTransformers2021} that reach state-of-the-art-results for Czech NLP. By comparing them, we can investigate the potential advantages of textual dependency in tasks and the benefits of in-domain pre-training. Both models
leverage RoBERTa~\cite{liuRoBERTaRobustlyOptimized2019} architecture with 52k/50k vocabulary size and 126M/124M parameters. Training data differ with RobeCzech having more diverse and well-rounded corpora, encompassing contemporary written Czech, articles from newspapers and magazines, and entries from Czech Wikipedia. In contrast, Fernet-News utilizes a single-domain corpus, predominantly featuring Czech news articles and broadcast transcripts. We hypothesize that the similarity of task and Fernet-News domains would lead to better task performance.

We fine-tuned both models for 2 epochs with linear decay, 0.1 warmup, 48 effective batch size, and AdamW as the optimizer. All layers were unfrozen except the embedding layer. Learning rates were selected based on the best validation score of a grid search over a 0.4 fraction of the training data. Possible learning rate values were:
\begin{itemize}
    \item RobeCzech: 3e-5, 4.5e-5, 7.5e-5
    \item Fernet-News: 1e-5, 2e-5, 3e-5
\end{itemize}
The proposed learning rate values for RobeCzech and Fernet-News differ due to the divergence of Fernet-News with higher learning rates. To deal with long texts, we chose to truncate them to the first 510 tokens.

\subsubsection{Final model.}
We additionally trained a RobeCzech with several enhancements. We first further pre-trained RobeCzech on the content of the article in FULL-SENTENCES setting~\cite{liuRoBERTaRobustlyOptimized2019} with a batch size of 192 and learning rate of 5e-5 for 10 epochs. We then trained the resulting model with a similar setting as in \ref{sec:base}.
We shortened the warmup to just 0.01 and fully removed it from the classifier. Finally, we added custom sampling with higher probabilities assigned to more recent articles.
\subsubsection{Results.}
\begin{table}[t]
    \centering
    \caption{Tasks F1 Macro scores for selected models on the Test set. We use \textbf{---} to denote that the model failed to converge for all tested learning rates.}
    \label{tab:results-test}
    \begin{tabular}{l c c c c}
        \toprule
        Model    & Source         & Category & Gender         & Day of week            \\
        \midrule
        Logistic Regression~~~ & 37.27          & 32.77    & 44.06          & 18.34                  \\
        RobeCzech (large data)  & 69.74         & 54.35    & 51.18          & 29.43                  \\
        Fernet-News (large data) & 69.39         & 53.97    & ---              & 29.24                  \\
        \midrule
        RobeCzech (small data) &           59.48 &           36.55 &           44.97 &           17.42 \\
        Fernet-News (small data) &               --- &           37.84 &               --- &           17.68 \\

        \midrule
        Final (large data)    & \textbf{71.04}         & \textbf{56.06} & \textbf{51.94} & \textbf{29.68} \\
        \bottomrule
    \end{tabular}
\end{table}
Table~\ref{tab:results-test} shows a significant improvement in Transformer models over Logistic Regression across the tasks. This demonstrates the importance of capturing textual dependencies for better performance.

Contrary to the initial expectation that Fernet-News would achieve higher scores due to its same-domain training data, RobeCzech outperformed Fernet-News across the tasks. One possible explanation could be RobeCzech's slightly higher capacity, which may be more important for long training. 

The Final model further improved the performance on all the tasks showcasing the importance of further in-domain pretraining and recency sampling.

\subsection{Small Dataset Setup}
In this setting, we evaluate a less-resourced scenario and assume we only have 50k most-recent training
instances with all task labels. Unlike the previous setup, here, we also include fine-tuning of the GPT-3 model, which is known to perform well in scenarios beyond English.
The evaluation is conducted on a smaller test set, a 10k subsample called \textbf{Test Small}.
\subsubsection{Pre-trained encoders.}
Both RobeCzech and Fernet-News were trained in this setting with the same parameters as in \ref{sec:base}.

\subsubsection{GPT-3.}
We selected the Ada variant from the GPT-3 family for our experiments. The model was trained in a multi-task setting, utilizing the article text (query) and corresponding task labels in Czech (text completion) as input.
The model was fine-tuned for two epochs.

\subsubsection{Results.}
\begin{table}[t]
    \centering
    \caption{Tasks F1 Macro scores for selected models on Test Small. We use \textbf{---} to denote that the model failed to converge for all tested learning rates.}
    \begin{tabular}{l c c c c}
        \toprule
        Model   & Source         & Category       & Gender         & Day of week                                                 \\
        \midrule
        RobeCzech (small data) & 75.12          & 37.88          & 47.45          & 17.41                                                       \\
        Fernet-News (small data)~~~ & ---              & 39.31          & ---              & 17.68                                                       \\
        GPT-3   & 67.30          & 44.76          & 42.92          & 19.49 \\
        \midrule
        RobeCzech (large data)  & \textbf{78.43} & \textbf{56.17} & \textbf{52.38} & \textbf{27.96}
        \\
        Fernet-News (large data) &           78.04 &           55.51 &               - &           27.25 \\
        \bottomrule
    \end{tabular}
    \label{tab:results-small}
\end{table}

Table~\ref{tab:results-small} shows the anticipated benefits of same-domain pre-training for Fernet-News became evident. When it converged, Fernet-News outperformed RobeCzech. Regarding GPT-3, it demonstrated its multilingual capabilities by outperforming both short models on two tasks.

For comparison, we also show the results of the fully trained RobeCzech and Fernet-News. Considering that the models in the small dataset setup were trained on less than 6\% of the data, the results can be regarded as fairly good, yet it is apparent that further training is beneficial.

We also observed a higher source task performance of the models on the set. The reason for this remains unclear, especially considering that the distributions are relatively similar across the test sets. The only noticeable change is iDnes.cz having slightly higher representation at the expense of Deník.cz.

\begin{table}[t]
        \centering
        \caption{Tasks F1 Macro scores for selected models on Test Human.}
        \label{tab:results-human}
        \begin{tabular}{l c c c c}
             \toprule
            Model    & Source & Category & Gender & Day of week                                                                     \\
            \midrule
            Human & 27.03               & 40.26               & 50.09          & 13.53\      \\
            Final (large data) & \textbf{71.22}      & \textbf{52.04}  &  \textbf{52.79} & \textbf{28.37} \\
            \bottomrule
        \end{tabular}       
\end{table}
\subsection{Human comparison}
\label{human-comparison}

Our findings revealed that the model outperforms human performance on every task, with the most significant improvement exceeding 44\% in the source task, as indicated in Table~\ref{tab:results-human}. 
We also evaluated inter-annotator agreement using Cohen's kappa, discovering that only the Category task exhibited significant agreement. The averaged scores were as follows: 0.08 for Source, 0.65 for Category, 0.20 for Gender, and 0.01 for Day Of Week. The low agreement and F1 macro scores observed for humans indicate that the tasks are indeed challenging.

\section{Related Work}

Benchmarking is the main way in which progress is measured in NLP. For English, aggregated benchmarks such as GLUE \cite{wang-etal-2018-glue} or SuperGLUE \cite{superglue} are often used to track the progress of pre-trained language models. The benchmarks cover various tasks that test language understanding, from sentiment analysis to challenging Winograd schemes. Multilingual benchmarks such as XTREME \cite{hu2020xtreme} cover Czech in a few tasks; however, they focus on cross-lingual transfer from English.

Czech-specific tasks span a variety of domains, such as Machine Translation~\cite{wmt22}, Question Answering~\cite{sqad}, Text Summarization~\cite{strakaSumeCzechLargeCzech2018a}, Sentiment Analysis~\cite{habernal-etal-2013-sentiment}, Named Entity Recognition~\cite{ner}, and Topic Classification tasks~\cite{kral-lenc-2018-czech},~\cite{leheckaComparisonCzechTransformers2021}.

Our research shares similarities with \cite{kral-lenc-2018-czech} and \cite{leheckaComparisonCzechTransformers2021}. Beyond news categories, we incorporate additional classification tasks into our study. Furthermore, our dataset is substantially larger and, owing to the variety of metadata gathered offers opportunities for additional tasks in the future.

\section{Conclusions}

We present a new dataset of Czech news articles that covers news stories between 2000 and 2022. We defined four classification tasks: news source, news category, inferred gender of the author, and day of the week of publishing the paper. Manual annotation of a part of the dataset showed that the tasks are challenging for humans.

The classification results are achieved by fine-tuning Czech pre-trained encoder models. Despite the recent development in pre-trained generative language models, pre-trained encoders outperform GPT-3 in two of four tasks. The best model largely outperforms human guesses, except for the inferred gender classification. The generally low performance on the inferred gender classification suggests that the risk that the pre-trained model would discriminate because of implicitly assuming authors' gender is probably low.

\section{Acknowledgment}

We thank Jindřich Helcl for comments on the early draft of the paper. The work on this paper was funded by the PRIMUS/23/SCI/023 project of Charles University and has been using resources provided by the LINDAT/CLARIAH-CZ Research Infrastructure (https://lindat.cz), supported by the Ministry of Education, Youth and Sports of the Czech Republic (Project No. LM2023062).

%
%
%
\bibliographystyle{splncs04}
\bibliography{bibliography}

\end{document}